\documentclass[10pt,twocolumn,letterpaper]{article}

\usepackage{cvpr}
\usepackage{times}
\usepackage{epsfig}
\usepackage{graphicx}
\usepackage{amsmath}
\usepackage{amssymb}
\usepackage{nicefrac} 
\usepackage{booktabs}     
\usepackage{algorithm,algorithmic}
\usepackage[caption=false,font=normalsize,labelfont=sf,textfont=sf]{subfig}
\usepackage[pagebackref=true,breaklinks=true,letterpaper=true,colorlinks,bookmarks=false]{hyperref}

 \cvprfinalcopy % *** Uncomment this line for the final submission

\def\cvprPaperID{7468} % *** Enter the CVPR Paper ID here

\ifcvprfinal\fi

\begin{document}

%%%%%%%%% TITLE
\title{Accelerating Optimization Algorithms With Dynamic Parameter Selections Using Convolutional Neural Networks For Inverse Problems In Image Processing}

\author{Byung~Hyun~Lee \qquad \qquad Se~Young~Chun\thanks{Corresponding to {\tt sychun@unist.ac.kr}}\\
School of Electrical and Computer Engineering, UNIST, Republic of Korea
}

\maketitle
\thispagestyle{empty}

%%%%%%%%% ABSTRACT
\begin{abstract}
Recent advances using deep neural networks (DNNs) for solving inverse problems in image processing have
significantly outperformed conventional optimization algorithm based methods.
Most works train DNNs to learn 
1) forward models and image priors implicitly for direct mappings from given measurements to solutions,
2) data-driven priors as proximal operators in conventional iterative algorithms, or
3) forward models, priors and/or static stepsizes in unfolded structures of optimization iterations.
Here we investigate another way of utilizing convolutional neural network (CNN) for empirically 
accelerating conventional optimization for solving inverse problems in image processing.
We propose a CNN to yield parameters in optimization algorithms 
that have been chosen heuristically, 
but have shown to be crucial for good empirical performance.
Our CNN-incorporated scaled gradient projection methods, without compromising theoretical properties, 
significantly improve empirical convergence rate over 
conventional optimization based methods 
in large-scale inverse problems such as image inpainting, 
compressive image recovery with partial Fourier samples, 
deblurring and sparse view CT. 
During testing, our proposed methods dynamically select parameters every iterations 
to speed up convergence robustly for different degradation levels, noise, or regularization parameters 
as compared to direct mapping approach.
\end{abstract}

%%%%%%%%% BODY TEXT
\section{Introduction}

Optimization based solvers for inverse problems have been widely investigated for image processing applications such as denoising~\cite{Mairal:2009gl}, inpainting~\cite{Roth:2005hu}, deblurring~\cite{Zoran:2011jn}, image recovery from incomplete Fourier samples~\cite{Patel:2012gz}, and image reconstruction from noisy Radon transformed measurements~\cite{Zhong:2013in}. 
A typical pipeline for them is to construct an objective function with accurate forward modeling of image degradation processes and with reasonable image priors such as minimum total variation (TV) and/or sparsity in wavelet (or learned transformed) domain, 
and then to optimize the objective function to yield a solution, a recovered image, using theoretically well-grounded optimization algorithms such as  iterative shrinkage-thresholding algorithm (ISTA)~\cite{figueiredo2003algorithm}, fast ISTA (FISTA)~\cite{Beck:2009tk}, approximate message passing (AMP) based algorithm~\cite{donoho2009message}, or
alternating directional method of multipliers (ADMM)~\cite{boyd2011distributed}.
Many related works have been proposed to improve theoretical convergence rates of algorithms and/or to design good image priors to regularize ill-posed inverse problems.
		
Deep neural networks (DNNs) have revolutionized the ways of solving inverse problems in image processing. 
These methods have significantly outperformed conventional optimization based methods 
by significantly improving the image quality of solutions and computation speed.
There are largely three ways of using DNNs for inverse problems in image processing:
1) direct mapping DNNs from measurements (or analytic reconstructions) to solutions 
by implicitly learning forward models and image priors~\cite{xie2012image,zhang2017beyond,Lim:2017it,Kulkarni:2016jea,Jin:2017iz},
2) DNN based proximal operators for iterative algorithms by explicitly learning image priors~\cite{Chang:2017fz,metzler2017learned,tirer2018image,he19tmi,ryu19a}, or
3) unfolded structure DNNs inspired by conventional optimization iterations by learning forward models, priors and/or stepsizes 
in optimizations~\cite{gregor2010learning,sun2016deep,Moreau:2017wj,Giryes:2018by,Zhang:2018wz,Chen:2018vc,Liu:2019ta}.
Recently, it has been theoretically shown that unfolded LISTA has asymptotic linear convergence property for compressive sensing (CS) recovery problems~\cite{Chen:2018vc,Liu:2019ta}.
However, these methods have been applied only to small-scale CS recovery problems due to large number of parameters 
to train or to determine. 

Most previous works determine static stepsizes with Lipschitz constant / training processes or dynamic stepsizes with backtracking.
However, we argue that for empirically fast convergence rate, 
stepsizes in optimization iterations must be determined dynamically for different problems every iterations.
Here we investigate an alternative way of utilizing CNNs for empirically accelerating optimization for solving large-scale inverse problems in image processing.
We propose CNNs estimating a near optimal stepsize (or a diagonal matrix) per iteration for given problems
to accelerate empirical convergence rates of conventional optimizations.
These parameters in optimization algorithms have been selected heuristically, but have shown to be crucial for good empirical performance.
Our CNN-incorporated scaled gradient projection (SGP) methods, without compromising theoretical properties, 
significantly improve empirical convergence rate over 
conventional optimization based methods such as  ISTA~\cite{figueiredo2003algorithm} / FISTA~\cite{Beck:2009tk} with backtracking
in large-scale inverse problems such as image inpainting, 
CS image recovery with partial Fourier samples, image deblurring and sparse view CT reconstruction.
Our ways of using DNN select parameters every iteration to speed up empirical convergence robustly
for different degradation levels, noise, or regularization parameters 
as compared to typical direct mapping CNN ($e.g.$, U-Net~\cite{Ronneberger:2015vw}).

Here are our contributions: We
1) propose small CNNs to dynamically determine stepsizes in optimizations every iterations, 
2) propose CNN-incorporated SGP methods without compromising convergence properties, and 
3) demonstrate the performance and robustness of our methods for large-scale inverse problems in image processing.

\section{Related Works}

DNN-based direct mapping approaches 
have yielded state-of-the-art performance in image quality and computation speed for inverse problems in image processing such as 
 image inpainting~\cite{xie2012image}, image denoising~\cite{zhang2017beyond}, single image super resolution~\cite{Lim:2017it}, sparse image recovery~\cite{Kulkarni:2016jea} and medical image reconstruction~\cite{Jin:2017iz}. 
However, they also have limitations such as no mechanism to ensure 
that a solution is corresponding to a given measurement through forward models by correcting for intermediate errors in a current solution.
Moreover, DNN-based direct mapping methods are based on black-box models with limited interpretability on the solutions for inverse problems.
In contrast, conventional optimization based approaches often have theoretical guarantees for the exact recovery~\cite{Candes:2006eq} 
or have interpretable converged solutions such as K-sparse images in wavelet domain with minimum distances to given measurements through forward models.

DNNs inspired by unfolded optimization iterations were also investigated with finite number of iterations.
Learned ISTA (LISTA)~\cite{gregor2010learning} was the first work of this type to propose DNNs that implicitly 
learn forward models, image priors and stepsizes in optimizations from data. 
LISTA has been extended to unfolded LISTA with learned weights~\cite{Chen:2018vc} 
and ALISTA with analytically determined weights~\cite{Liu:2019ta}.
The original LISTA compromised ISTA's theoretical convergence properties, but there have been efforts to understand 
convergence properties in unfolded structure of DNNs~\cite{Moreau:2017wj,Giryes:2018by}.
Recently, it has been theoretically shown that unfolded LISTA has asymptotic linear convergence property 
for CS recovery problems~\cite{Chen:2018vc}.
However, these methods have been applied to small-scale CS recovery problems 
due to large number of parameters to determine.
If image size is 256$\times$256 and the compression ratio is about 20\%,
LISTA~\cite{gregor2010learning} and LISTA-CPSS~\cite{Chen:2018vc} require about 4,465M, 2,727M parameters to train, 
while our proposed U-Net based CNN requires only 7M parameters for 256$\times$256 images.
Recently, ALISTA~\cite{Liu:2019ta} proposed methods for determining analytical weights using dictionary learning or
convolutional sparse coding, but it does not seem to work for general forward models yet.

ADMM-net~\cite{sun2016deep} and ISTA-Net~\cite{Zhang:2018wz} are also based on unfolded optimization iterations of ADMM and ISTA, respectively, but unlike LISTA approaches,
they utilized forward models in their networks and trained convolutional neural networks (CNNs) for image priors such as transformations, parametrized non-linear functions
and for optimization parameters such as stepsize. 
Using forward models explicitly allows these methods to deal with large-scale inverse problems.
Similarly, there have been works using DNNs only for proximal operators in iterative algorithms~\cite{Chang:2017fz,metzler2017learned,tirer2018image,he19tmi,ryu19a}.
Unlike ADMM-net and ISTA-Net with fixed number of iterations, these methods have flexibility of running any number of iterations for different cases.
However, they focused on using DNNs as image priors within conventional optimization framework, 
rather than investigating acceleration of convergence. 
Most works in this category have static stepsizes that have been determined heuristically, selected conservatively to ensure convergence (e.g., 1/Lipschitz constant), or learned from data.

Lastly, there have been a few recent attempts to learn optimization algorithms as DNN based functions of gradients~\cite{andrychowicz2016learning} and as policies of selecting algorithms using reinforcement learning~\cite{Li:2017lto}. These are similar to our proposed methods in dynamically determining algorithms every iterations. However, our methods are within the framework of SGP methods with theoretical convergence properties.

\section{Background}

\subsection{Proximal gradient method}

Consider an optimization problem of the form
\begin{equation}
	\underset{\mathbf{x}} {\text{min}} \, F(\mathbf{x}) = f(\mathbf{x}) + g(\mathbf{x})
	\label{eq:prob}
\end{equation}
where $f$ is convex, differentiable and $g$ is convex, subdifferentiable for $\mathbf{x}$. Then,
the following update equation at the $k$th iteration is called the proximal gradient method (PGM):
\begin{equation}
	\mathbf{x}^{(k+1)} = \text{prox}_{t_k}(g)(\mathbf{x}^{(k)}-t_k{\nabla}f(\mathbf{x}^{(k)})),
\end{equation}
where 
\(
	\text{prox}_t(g)(\mathbf{x}) = \arg \min_{\mathbf{z}} g(\mathbf{z}) + \nicefrac1{2t}\| \mathbf{z} - \mathbf{x}\|^2
\)
and $\| \cdot \|$ denotes $l_2$-norm.
PGM guarantees the convergence of $F(\mathbf{x}^{(k)})$ to $F^*$ at the solution $\mathbf{x}^*$ with the rate of $O(1/k)$. 

One way to determine $t_k$ is based on a majorization-minimization technique for the cost function $F$ and its quadratic surrogate function for each iteration. 
${\nabla}f$ is usually assumed to be Lipschitz continuous on a given domain and the reciprocal of a Lipschitz constant of ${\nabla}f$ is used for $t_k$. Another popular method to choose $t_k$ is a backtracking method. 
Note that both methods do not seek the largest possible stepsize since it is often more efficient to calculate the next iteration with conservative sub-optimal stepsize than to perform time-consuming stepsize optimization. Thus, if there is a way to quickly calculate near optimal stepsizes, it could help to accelerate empirical convergence rates.

\subsection{Scaled gradient projection method}
\label{sec:sgp}

Recently, scaled gradient projection (SGP) methods with the convergence rate $O(1/k)$
have been proposed and have empirically demonstrated their general convergence speed improvements over FISTA with $O(1/k^{2})$~\cite{Bonettini:2008ud,Bonettini:2015hu}.
The problem (\ref{eq:prob}) can be seen as a constrained optimization problem
\begin{equation} 
	\underset{\mathbf{x} \in S}{\text{min}} \, f(\mathbf{x})
\label{eq:constrain}
\end{equation}
where $S = \{\mathbf{x} \in \mathbb{R}^n: g(\mathbf{x}) \leqslant \epsilon \}$ that is a convex set due to the convexity of $g$ and $\epsilon$ is determined by $g$. Then, an iterative algorithm can be formulated as the PGM given by
 \begin{equation}
\mathbf{y}^{(k)} = \mathbf{P}_{S}(\mathbf{x}^{(k)}-t_k{\nabla}f(\mathbf{x}^{(k)}))
 \end{equation}
 where
\(
 \mathbf{P}_{S}(\mathbf{x})= \arg \min_{\mathbf{z} \in S} \|\mathbf{z}-\mathbf{x}\|^2.
 \)
Whenever $\mathbf{y}^{(k)}\neq\mathbf{x}^{(k)}$, $\mathbf{y}^{(k)}-\mathbf{x}^{(k)}$ is a descent direction at $\mathbf{x}^{(k)}$ for the problem (\ref{eq:constrain}) and thus 
its inner product with ${\nabla}f(\mathbf{x}^{(k)})$ becomes negative.
$\mathbf{y}^{(k)}=\mathbf{x}^{(k)}$ implies that $\mathbf{x}^{(k)}$ is a stationary point. 
Since $\mathbf{y}^{(k)}-\mathbf{x}^{(k)}$ is a descent direction at $\mathbf{x}^{(k)}$, 
Armijo line search can generate a convergent sequence of $\{\mathbf{x}^{(k)}\}_{k=0} ^{\infty}$ that satisfy Armijo condition~\cite{Armijo:1966hc}:
  \begin{equation}
 	f(\mathbf{x}^{(k)} + \gamma^{(k)} \mathbf{z}^{(k)}) \leqslant f(\mathbf{x}^{(k)}) + \beta \gamma^{(k)} {\nabla}f(\mathbf{x}^{(k)})^T\mathbf{z}^{(k)}
\label{eq:armijo}
 \end{equation} 
where $\mathbf{z}^{(k)} = \mathbf{y}^{(k)}-\mathbf{x}^{(k)}$ and $\beta, \gamma^{(k)} \in (0, 1]$ for $\forall k$.

In the problem (\ref{eq:constrain}), SGP methods introduced an additional symmetric positive definite matrix $D_k$ in front of ${\nabla}f(\mathbf{x}^{(k)})$. Symmetry and positive definiteness are necessary conditions for a Hessian matrix of $f$ for Newton's method and they are also important conditions for quasi-Newton methods. Newton-type methods usually converge with fewer iterations than first-order optimization methods, but they are computationally demanding especially for 
large-scale input data.
SGP methods exploit symmetry and positive definiteness with the aim of less 
computational burdens 
while they can refine a direction vector $\mathbf{d}^{(k)}$ of the diagonal elements in $D_k$
to accelerate convergence rate. SGP methods are based on Armijo line search since $\mathbf{d}^{(k)}$ remains as a descent direction 
with the conditions on $D_k$ at $\mathbf{x}^{(k)}$. They are also applicable for proximal operators.

For the convergence of SGP methods, it requires additional condition for $D_k$. Define $\mathcal{D}_\delta$ for $\delta \geqslant 1$ as the set of all symmetric positive definite matrices whose eigenvalues are in the interval $[1/\delta, \delta]$. Then, for $\delta_k$ such that $D_k \in \mathcal{D}_{\delta_k}$, the condition $\Sigma_{k=0}^\infty (\delta_k^2 -1) < \infty$, $\delta_k \geqslant 1$ should be satisfied.   
$\lim_{k\rightarrow\infty} D_k$ becomes an identity matrix and an iteration becomes similar to PGM. 
Appropriate $D_k$~\cite{Bonettini:2015hu} accelerated empirical convergence over fast PGMs such as FISTA.

For a given proximal operator
\(
	\text{prox}_{D}(g)(\mathbf{x}) = \arg \min_{\mathbf{z}} g(\mathbf{z}) + \nicefrac{1}{2}(\mathbf{z}-\mathbf{x})^TD^{-1}(\mathbf{z}-\mathbf{x}),
\)
the SGP method is summarized in Algorithm~\ref{alg:spg}.
Finding $\{D_k\}_{k=0}^\infty$ that can accelerate convergence still remains as an open problem. 
We propose to replace these heuristic decisions with DNNs.

\begin{algorithm}[!t]
\caption{Scaled Gradient Projection (SGP)~\cite{Bonettini:2008ud,Bonettini:2015hu}}
\label{alg:spg}
\begin{algorithmic}
	\STATE Given $0<t_{min} \leqslant t_{max}$, $\delta \geqslant 1$, and  $\beta,\eta \in (0, 1) $,
	\FOR {$k = 0, 1, 2, ..., K$ }
		\STATE Set $\gamma^{(k)} \leftarrow 1$
		\STATE Choose $t_k \in [t_{min}, t_{max}]$, $0 < \delta_k < \delta$ and $D_k \in \mathcal{D}_{\delta_k}$
		\STATE $\mathbf{z}^{(k)} = \text{prox}_{t_k D_k} (g) (\mathbf{x}^{(k)}- t_k D_k {\nabla}f(\mathbf{x}^{(k)})) - \mathbf{x}^{(k)}$
		\WHILE {\text{(\ref{eq:armijo}) is not satisfied}} 
			\STATE $\gamma^{(k)} \leftarrow \eta \, \gamma^{(k)}$
		\ENDWHILE
		\STATE $\mathbf{x}^{(k+1)} = \mathbf{x}^{(k)}+\gamma^{(k)}\mathbf{z}^{(k)}$
	\ENDFOR
\end{algorithmic}
\end{algorithm}

\section{Learning-based stepsize selection}
\label{sec:learnedstep}

We conjecture that DNN can be trained to generate a near-optimal stepsize per iteration if a current estimate and the gradient of a cost function at that estimate are given.
Since no ground truth is available for the optimal sequence of stepsizes over all iterations,
we propose to train a stepsize DNN to yield a greedy near-optimal stepsize per iteration by
minimizing the distance between the estimated vector at the next iteration and 
the converged solution at each iteration.

\subsection{Learning a stepsize for an iteration}
\label{sec:onestep}

To learn stepsizes by a DNN, a set of $N$ solution vectors $\{\mathbf{x}_*^n\}_{n=1}^N$ of optimization problems was generated and used as ground truth data.
Solution vectors can be obtained by optimizing the original problems using any convex optimization algorithm ($e.g.$, FISTA with 1200 iterations).
Suppose that the estimates at the $k$th iteration form a set of training data or 
$\{\mathbf{x}^{(k), n}\}_{n = 1}^N$ that will be fed into the DNN.
We denote the output of the DNN as a set of positive real numbers $\{t_{k, n}\}_{n=1}^N$ for stepsizes. 
Then, 
a set of vectors  $\{\tilde{\mathbf{x}}^{(k), n}\}_{n = 1}^N$ can be obtained at the next iteration: % such that
\begin{equation}
	\tilde{\mathbf{x}}^{(k), n} = \mathcal{T}_{\lambda t}\left( \mathbf{x}^{(k), n} - t_{k,n}{\nabla}f(\mathbf{x}^{(k), n}) \right)
	\label{eq:method1}
\end{equation}
where
\(
\mathcal{T}_{\lambda t}(\mathbf{x})_i = \max\{|x_i|-{\lambda t}, 0\} \text{sgn}(x_i)
\)
for the $i$th element in the vector (soft threshold).

The desired stepsizes $\{t_{k, n}\}_{n=1}^N$ for the $n$th images at the $k$th iteration can be obtained by training the DNN $\Psi( \cdot ;\bf \Theta)$ to 
minimize the following loss function with respect to $\bf \Theta$:
\begin{align}
\underset{\bf \Theta}{\text{min}} \, \frac1{2} \left\| \mathbf{x}_*^n - \mathcal{T}_{\lambda t} \left( \mathbf{x}^{(k), n} - 
t_{k, n} (\bf \Theta)  {\nabla}f(\mathbf{x}^{(k), n}) \right) \right\|^2
\end{align}
where $t_{k, n} (\bf \Theta) = \Psi \left( \mathbf{x}^{(k), n}, {\nabla}f(\mathbf{x}^{(k), n}) ;\bf \Theta \right) $
and then by evaluating 
\(
t_{k, n} = \Psi \left( \mathbf{x}^{(k), n}, {\nabla}f(\mathbf{x}^{(k), n}) ;\bf \Theta \right).
\)
After $\{\tilde{\mathbf{x}}^{(k), n}\}_{n=1}^N$ are evaluated by the learned stepsizes $\{t_{k, n}\}_{n=1}^N$ using (\ref{eq:method1}), 
we propose to generate another set of vectors for the next iteration $\{\mathbf{x}^{(k+1), n}\}_{n=1}^N$ by using a conventional stepsize based on
Lipschitz constant $L$:
\begin{equation}
	\mathbf{x}^{(k+1), n} = \mathcal{T}_{{\lambda}/{L}} \left(\tilde{\mathbf{x}}^{(k), n} - t_L {\nabla}f(\tilde{\mathbf{x}}^{(k), n} \right)
\end{equation}
where $t_L = 1 / L$. 
This additional step was necessary since $\{\tilde{\mathbf{x}}^{(k), n}\}_{n=1}^N$ were not often improved over $\{\mathbf{x}^{(k), n}\}_{n = 1}^N$ when the DNN training was not done yet.

To sum, one iteration of our proposed method consists of two steps: 1) the first operation moves a current estimate towards its solution,
2) the second operation is applied for initial training of DNNs.
In our simulations, our proposed training method worked well to reduce the loss quickly.
 
The same training method with a diagonal matrix can be applied by replacing the stepsize $t_{k,n}$ in (\ref{eq:method1}) with a diagonal matrix $D_{k, n}$. 
The output dimension of the DNN  $\Psi( \cdot ;\bf \Theta)$ must be changed from $1$ to $G$ with its backpropagations.

\subsection{Learning stepsizes for further iterations}

%Based on one iteration training, 
Now we propose to further train DNNs to generate stepsizes for multiple 
iterations. Inspired the training strategy in~\cite{gupta2018tmi}, we define the following cumulative loss function:
 \begin{equation}
 	\mathcal{L}_k = \textstyle \sum_{i=0}^k \, \tilde{\mathcal{L}}_i
	\label{eq:loss1}
 \end{equation}
where 
\(
	\tilde{\mathcal{L}}_k
	 = \nicefrac1{2} {\sum_{n=1}^N} \|\mathbf{x}_*^n - \tilde{\mathbf{x}}^{(k), n} \|^2,
\)
$\tilde{\mathbf{x}}^{(k), n}$ is defined in (\ref{eq:method1}), and 
new input datasets as well as ground truth labels are defined as 
\(
 	\mathcal{I}_k = \cup_{i=1}^k \{ \mathbf{x}^{(i), n}, {\nabla}f(\mathbf{x}^{(i), n}) \}_{n=1}^N
\)
and
\(
	\mathcal{O}_k = \cup_{i=1}^k  \{\mathbf{x}_*^n\}_{n=1}^N
\)
where $\mathcal{O}_k$ contains duplicated sets.
 
Suppose that the DNN is to learn stepsizes of the first $K$ iterations. 
Initially, the DNN is trained with the input data set $\mathcal{I}_0$ and the ground truth label $\mathcal{O}_0$ at the $0$th iteration using the procedure in Section~\ref{sec:onestep}. Then, in the next iteration, the DNN is re-trained with the input data set $\mathcal{I}_1$ and the ground truth label $\mathcal{O}_1$ at the first iteration. This training process is repeated $K+1$ times so that the DNN can be trained cumulatively as summarized in Algorithm~\ref{alg:mult}.

\begin{algorithm} [!b]
\caption{Stepsize Learning for Multiple Iterations}
\label{alg:mult}
\begin{algorithmic}
	\STATE Given $\mathcal{I}_k$, $\mathcal{O}_k$,
	\FOR {$k = 0, 1, 2, ..., K$ }
		\STATE Train DNN with the input set $\mathcal{I}_k$ and the label set $\mathcal{O}_k$
	\ENDFOR
\end{algorithmic}
\end{algorithm}

We expect that our trained DNN should yield near-optimal stepsizes for the first $K$ iterations, but may not be able to yield good stepsizes in later iterations that are larger than $K$. Thus, $K$ should be selected based on the trade-off between image quality and computation time.

\section{DNN-incorporated convergent algorithms}

\subsection{SGP method as framework}

The SGP method is described in Algorithm~\ref{alg:spg}. 
If $D_k$ is an identity matrix and $\gamma^{(k)}$ is equal to 1 for $\forall k$, the SGP method is reduced to the PGM. 
Thus, the SGP method is a generalized version of the PGM by additionally multiplying a symmetric positive definite matrix with the gradient of a loss function that guarantees $\mathbf{z}^{(k)}$ to be a descent direction and by enforcing the Armijo condition for convergence. 
However, there is no known method to determine $\{D_k\}_{k=0}^\infty$ that can accelerate and guarantee convergence. We propose DNN to determine
$\{D_k\}_{k=0}^\infty$ 
that can be trained using the learning procedure in Section~\ref{sec:learnedstep}.
Since it is also possible for the DNN to yield $\{D_k\}_{k=0}^\infty$ that may not satisfy necessary conditions, 
we proposed to relax the SGP method to selectively use DNN based stepsize (or diagonal matrix) estimation or conservative Lipschitz constant based stepsize
to guarantee convergence as summarized in Algorithm~\ref{alg:drs}. We call this the direction relaxation scheme (DRS).

\begin{algorithm}[!b]
\caption{Direction Relaxation Scheme (DRS)}
\label{alg:drs}
\begin{algorithmic}
	\STATE Given $\mathbf{x}^{(k)}$, ${\nabla}f(\mathbf{x}^{(k)})$, $\gamma^{(k)}_1, \alpha \in (0, 1]$, $useCNN \in \{0, 1\}$, Lipschitz constant $L$ of $f$, and a trained DNN function $\Psi$ 
	\IF {$useCNN$}
		\STATE $\mathbf{d}^{(k)} = \Psi (\mathbf{x}^{(k)}, {\nabla}f(\mathbf{x}^{(k)}))$ %, \qquad $\mathbf{d}^{(k)} = \mathrm{diag}(D_k)$
		\STATE $\tilde{\mathbf{x}}^{(k)} = \mathcal{T}_{\mathbf{d}^{(k)}}(\mathbf{x}^{(k)}-D_k{\nabla}f(\mathbf{x}^{(k)}))$
		\STATE $\mathbf{z^{(k)}_1} = \mathcal{T}_{t_{L}} (\tilde{\mathbf{x}}^{(k)}-t_{L}{\nabla}f(\tilde{\mathbf{x}}^{(k)})) - \mathbf{x}^{(k)}$
		\STATE $\mathbf{z}^{(k)}_2 = \mathcal{T}_{t_{L}} (\mathbf{x}^{(k)}-t_{L}{\nabla}f(\mathbf{x}^{(k)})) - \mathbf{x}^{(k)}$
		\STATE		
		\IF {$\gamma^{(k)}_1 ||\mathbf{z}^{(k)}_1|| > \alpha (1-\gamma^{(k)}_1) || \mathbf{z}^{(k)}_2 || $}
			\STATE $z^{(k)} = \gamma^{(k)}_1 \mathbf{z}^{(k)}_1 + (1- \gamma^{(k)}_1)\mathbf{z}^{(k)}_2$	
		\ELSE
			\STATE $\mathbf{z}^{(k)} = \mathbf{z}^{(k)}_2$, \quad
			$useCNN \leftarrow false$
		\ENDIF
	\ELSE
		\STATE 	$\mathbf{z}^{(k)} = \mathcal{T}_{t_{L}} (\mathbf{x}^{(k)}-t_{L}{\nabla}f(\mathbf{x}^{(k)})) - \mathbf{x}^{(k)}$
	\ENDIF
\end{algorithmic}
\end{algorithm}

\subsection{Proposed relaxation algorithms with DNN}

\begin{algorithm} [!t]
\caption{Proposed SGP Algorithm with Stepsize DNN}
\label{alg:proposed}
\begin{algorithmic}
	\STATE Given $\eta_1, \eta_2, \beta \in (0, 1)$, $\mathbf{x}^{(0)}$, $K \in \mathbb{N}$,
	$\gamma^{(0)}_1 = 1$, $useCNN \leftarrow true$
	\FOR {k = 0, 1, 2, ..., K}
		\STATE Generate $\mathbf{z}^{(k)}$ by DRS in Algorithm~\ref{alg:drs}
		\STATE Set $\gamma^{(k+1)}_1 \leftarrow \gamma^{(k)}_1$ and $\gamma^{(k)}_2 \leftarrow 1$
		\WHILE{(\ref{eq:armijo}) is not satisfied with $\gamma^{(k)}_2$}
			\STATE $\gamma^{(k+1)}_1 \leftarrow \eta_1 \gamma^{(k+1)}_1$, \quad
			$\gamma^{k}_2 \leftarrow \eta_2 \gamma^{(k)}_2$
		\ENDWHILE
		\STATE $\mathbf{x}^{(k+1)} = \mathbf{x}^{(k)}+\gamma^{(k)}_2 \mathbf{z}^{(k)}$
	\ENDFOR
\end{algorithmic}
\end{algorithm}

For the DRS in Algorithm~\ref{alg:drs},
note that $\text{prox}_{D}(g) = \mathcal{T}_{\mathbf{d}}$ when $g$ is $l_1$-norm and $D$ is a diagonal matrix whose diagonal elements form a vector $\mathbf{d}$.
$\mathbf{z}^{(k)}_1$ represents a wild search direction generated by the trained DNN and $\mathbf{z}^{(k)}_2$ is a conservative search direction from conventional
Lipschitz constant based stepsize. Then, depending on the relationship between $\mathbf{z}^{(k)}_1$ and $\mathbf{z}^{(k)}_2$, the final search direction
will be either a linear combination of both of them or $\mathbf{z}^{(k)}_2$ alone.
For the DNN to generate a single stepsize, $D_k$ will be an identity matrix multiplied by that stepsize.

We propose to incorporate the DNN based DRS method into the SGP algorithm as detailed in Algorithm~\ref{alg:proposed}.
$\mathbf{z}^{(k)}$ is a search direction to yield the estimated vector for the next iteration. As in Algorithm~\ref{alg:drs},
$\mathbf{z}^{(k)}$ is either the weighted average of $\mathbf{z}^{(k)}_1$ and $\mathbf{z}^{(k)}_2$ with $\gamma_1 ^{(k)}$ or $\mathbf{z}^{(k)}_2$ itself.
The value of $\gamma_1 ^{(k)}$ was initially set to be $1$ and it remains the same or decreases by a factor $\eta_1$ over iterations depending on the Armijo condition at each iteration $k$.
The ratio of the weight for $\mathbf{z}^{(k)}_1$ to the weight for $\mathbf{z}^{k}_2$ in $\mathbf{z}^{k}$ is evaluated at each iteration. 
Initially, $\mathbf{z}^{(k)}_1$ using the trained DNN is dominant in $\mathbf{z}^{(k)}$, but
$\mathbf{z}^{(k)} = \mathbf{z}^{(k)}_2$ for later iterations and the DNN will not be used eventually.
Thus, our proposed algorithm is initially the SGP with DNN search directions and becomes the PGM for later iterations.

Note that the proposed DRS method with relaxation only determines a search direction for the next estimate in a descent direction for inverse problems. 
$\gamma^{(k)}_2$ is the final stepsize parameter, starting from $1$ and decreases its value %until %determined starting from 1 and decreases its value 
by a factor $\eta_2$ until it satisfies the Armijo condition. 
Therefore, our proposed method in Algorithms~\ref{alg:drs} and \ref{alg:proposed} using the trained DNN is converging theoretically.

\section{Simulation results}

\subsection{Inverse problem settings}

We performed various image processing simulations such as image inpainting, CS image recovery with partial Fourier samples,
image deblurring and large-scale sparse-view CT
to evaluate our proposed methods. 
An optimization problem for inverse problems in image processing has the following form: %The optimization problems to be solved in the experiemets are of the form
\begin{equation}
	\mathbf{\hat{x}}^n = \arg \underset{\mathbf{x}^n}{{\min}} \, \frac1{2}||\mathbf{A}_n \mathbf{x}^n - \mathbf{y}^n ||^2 + \lambda||\mathbf{x}^n ||_1
	\label{eq:simmodel}
\end{equation}
where $n$ is an index for image, $\mathbf{A}_n$ is a matrix that describes an image degradation forward process,
$\mathbf{y}^n$ is a measurement vector and $\lambda$ is a regularization parameter to balance between data fidelity and image prior.
An image is modeled to be in wavelet domain (three level symlet-4) so that $\mathbf{x}^n$ is a wavelet coefficient vector for an image.
So, the linear operator $\mathbf{A}_n$ is a measurement matrix (different for image by image) with an inverse sparsifying transform $\mathbf{W}^T$. 
For normalized measurement matrices, their Lipschitz constants are less than $1$ and  
normalized gradients helped to yield better results.
BSDS500 dataset~\cite{amfm_pami2011} with $256\times256$ images was used for all simulations where 450 / 50 images were used for training / testing, respectively.
$\lambda$ was set to be 0.1. 

We implemented our stepsize (or diagonal matrix) DNN based on U-Net architecture~\cite{Ronneberger:2015vw} and modified FBPConvNet~\cite{Jin:2017iz} using MatConvNet on MATLAB. 
Note that the input and output for the DNN are in a sparsifying transform domain and %it turned out that
they have improved the overall performance of inverse problems.
$3 \times 3$ convolution filters are used for all convolutional layers and batch normalization and rectified linear unit (ReLU) were used after each convolution layer. $2 \times 2$ max pooling was applied in the first half of the DNN and deconvolution layers / skip connections were used for the second half of the DNN. We reduced the number of layers in the original FBPConvNet to lower computation time (7M params). For stepsize learning, one fully connected layer was added at the end of the DNN to generate a single number.
All simulations were run on an NVIDIA Titan X.

We compared our proposed SGP methods with stepsize DNN (called Step-learned) and diagonal matrix DNN (called Diag-learned) to conventional algorithms such as ISTA and FISTA with backtracking. We also compared our proposed methods with the U-Net that was trained to yield ground truth converged images for (\ref{eq:simmodel}) from input measurements.
We chose U-Net to compare since it has been shown to yield good results in various image processing problems
with large-scale inputs and with different forward models for images
including inpainting and compressive image recovery~\cite{noise2noise}.
Unfortunately, we were not able to compare ours with ReconNet~\cite{Kulkarni:2016jea} or LISTA-CPSS~\cite{Chen:2018vc} that were limited to small-scale CS image recovery problems with the image patch sizes of 
$33 \times 33$ and $16 \times 16$, respectively, and empirically one fixed forward model $\mathbf{A}_n = \mathbf{A}, \forall n$.

NMSE was used for evaluation criteria~\cite{Chen:2018vc} using
\(
10 \log_{10} ( \| \mathbf{W}^T (\hat{\mathbf{x}} - \mathbf{x}^\mathrm{gt} ) \|^2 / \| \mathbf{W}^T \mathbf{x}^\mathrm{gt} \|^2).
\)
Note that all DNNs were trained with converged solutions $\mathbf{x}^*$ while the evaluations were done with the oracle solutions
$\mathbf{x}^\mathrm{gt}$ (the original BSDS500 dataset) for better evaluating the robustness of all methods under significantly different forward models and additional
measurement noise. All DNNs for our Step-learned/Diag-learned methods as well as U-Net to yield solutions directly were trained for the case of recovering from
noiseless 50\% samples with different forward models for different training images. Then, all methods were tested on
noiseless 30\%, 50\%, and 70\% samples for test dataset to evaluate performance as well as robustness.
For image recovery with partial Fourier samples, noisy 50\% samples (Gaussian with standard deviation 5)
were also used for further evaluation. 
We also performed image deblurring for Gaussian blur kernel with $\sigma = 2$ and 
sparse-view CT reconstruction with 144 views using $512\times512$ CT images to show the feasibility on
other applications.

\subsection{Image inpainting}

\begin{figure} [!t]
	\centering  
	\subfloat[\scriptsize Input 30, 50, 70\% samplings for inpainting]{
	\includegraphics[width=0.33\linewidth,trim={0 2cm 0 0},clip ]{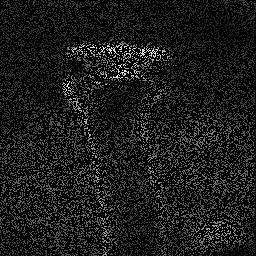}
	\includegraphics[width=0.33\linewidth,trim={0 2cm 0 0},clip ]{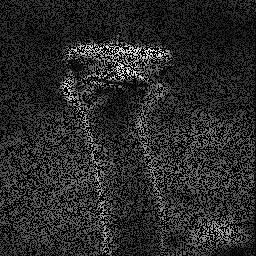}
	\includegraphics[width=0.33\linewidth,trim={0 2cm 0 0},clip ]{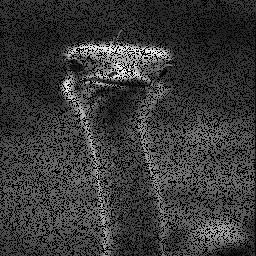}
	}
	\vfil 	\vspace{-0.5em}
	\subfloat[\scriptsize FISTA-b for 30, 50, 70\% samplings]{ 
	\includegraphics[width=0.33\linewidth,trim={0 2cm 0 0},clip ]{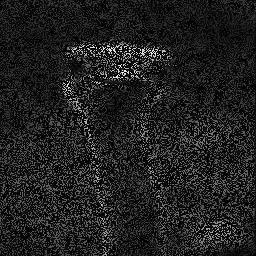}
	\includegraphics[width=0.33\linewidth,trim={0 2cm 0 0},clip ]{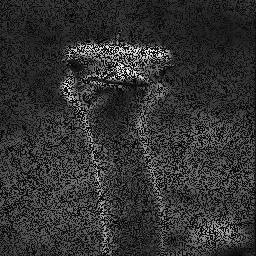}
	\includegraphics[width=0.33\linewidth,trim={0 2cm 0 0},clip ]{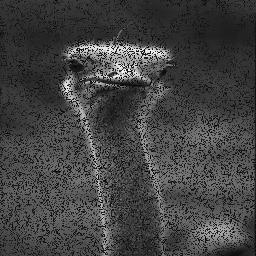}
	}
	\vfil 	\vspace{-0.5em}
	\subfloat[\scriptsize U-Net trained on 50\%, tested on 30, 50, 70\% samplings]{ 
	\includegraphics[width=0.33\linewidth,trim={0 2cm 0 0},clip ]{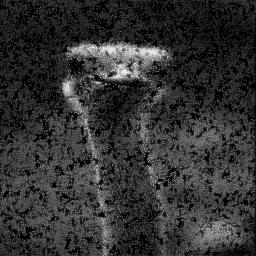}
	\includegraphics[width=0.33\linewidth,trim={0 2cm 0 0},clip ]{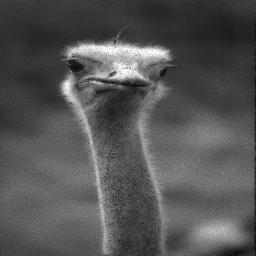}
	\includegraphics[width=0.33\linewidth,trim={0 2cm 0 0},clip ]{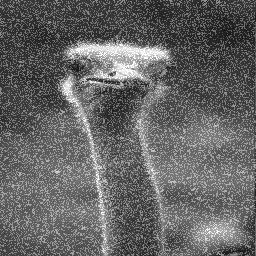}
	}
	\vfil 	\vspace{-0.5em}
	\subfloat[\scriptsize Step-learned SGP trained on 50\%, tested on 30, 50, 70\% samplings]{ 
	\includegraphics[width=0.33\linewidth,trim={0 2cm 0 0},clip ]{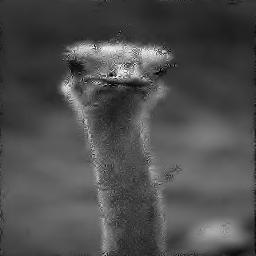}
	\includegraphics[width=0.33\linewidth,trim={0 2cm 0 0},clip ]{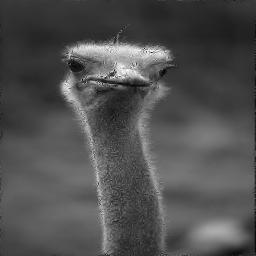}
	\includegraphics[width=0.33\linewidth,trim={0 2cm 0 0},clip ]{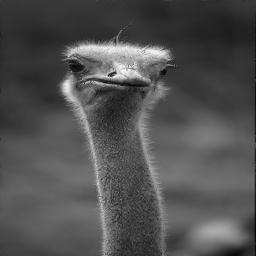}
	}
	\vfil 	\vspace{-0.5em}
	\subfloat[\scriptsize Diag-learned SGP trained on 50\%, tested on 30, 50, 70\% samplings]{ 
	\includegraphics[width=0.33\linewidth,trim={0 2cm 0 0},clip ]{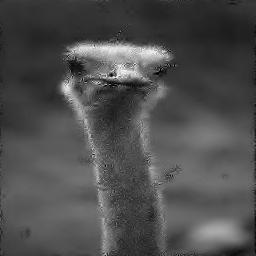}
	\includegraphics[width=0.33\linewidth,trim={0 2cm 0 0},clip ]{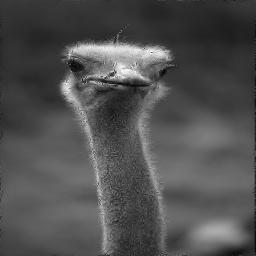}
	\includegraphics[width=0.33\linewidth,trim={0 2cm 0 0},clip ]{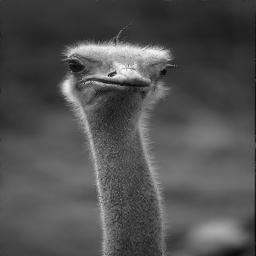}
	}
	\vfil 	\vspace{0.5em}
	\caption{Recovered images for inpainting using DNNs trained on 50\% and tested on 30, 50, 70\% samplings, respectively.
	FISTA-b, Step-learned and Diag-learned SGPs were run with 40 iterations.}
	\label{fig:inpaint}
	\vspace{-1em}
\end{figure}

\begin{figure} [!t]
	\centering  
	\subfloat[\scriptsize Trained on 50\%, Tested on 50\%]{\includegraphics[width=0.49\columnwidth]{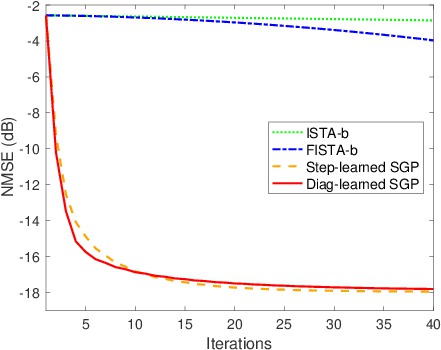}}
	\subfloat[\scriptsize Trained on 50\%, Tested on 30\%]{\includegraphics[width=0.49\columnwidth]{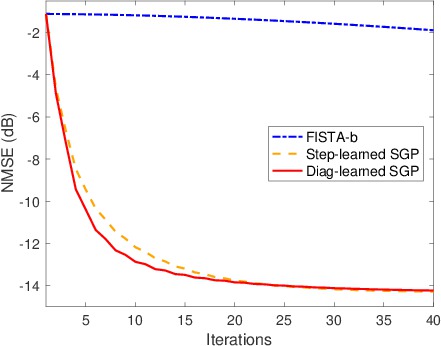}}
	\vspace{-0.5em}
	\vfil
	\subfloat[\scriptsize Trained on 50\%, Tested on 70\%]{\includegraphics[width=0.49\columnwidth]{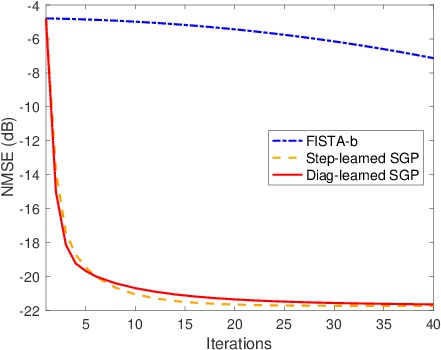}}
	\subfloat[\scriptsize Trained on $\lambda$, Tested on $\lambda/2$, $2\lambda$] {\includegraphics[width=0.49\linewidth,clip]{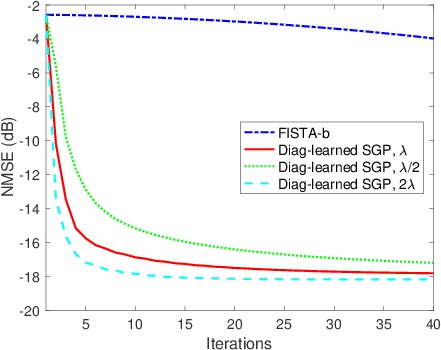}}
	\vspace{0.5em}
	\caption{NMSEs over iterations for the methods in inpainting trained on 50\% sampling with a regularization parameter  $\lambda$ and tested on
	50, 30, 70\% samplings (a-c) or with $\lambda/2$, $2\lambda$ (d).}
	\label{fig:inpaint_robust}
	\vspace{-1em}
\end{figure}

Our proposed methods were applied to inpainting problems with different sampling rates. 
$\mathbf{A}_n = \mathbf{M}_n \mathbf{W}^T$ was used where $\mathbf{M}_n$ is a sampling matrix that
is different for image. All DNNs were trained with 50\% samples (called U-Net-50\%, Step-learned-50\%, Diag-learned-50\%).
All results tested on different sampling rates are reported in Figures~\ref{fig:inpaint},~\ref{fig:inpaint_robust} and Table~\ref{tbl:inpaint}.

When all methods were tested on the same sampling rate (50\%), non-iterative U-Net instantly yielded the best image quality among
all methods including our proposed methods and even the ground truth converged images (FISTA at 1200 iteration yielded -18.15dB). 
Both of our proposed methods at 20 iteration yielded image qualities comparable to the converged images, while FISTA at 100 iteration performed much
worse than our proposed methods at 20 iteration.

However, U-Net did not show robust performance and yielded substantial artifacts 
for the tests with different sampling rates such as 30\% and 70\%. However, our proposed methods without re-training
yielded robust accelerations for different test cases as illustrated in Figure~\ref{fig:inpaint_robust}.

\begin{table}[!b]
  \caption{Averaged NMSE (dB) of all methods trained with 50\%, tested on various cases for inpainting.}
  \label{tbl:inpaint}
  \centering \footnotesize
  \begin{tabular}{c | ccc}
    \toprule
    Method     & Test-30\%     & Test-50\% & Test-70\% \\
    \midrule
    FISTA@100 & -5.20$\pm$1.01  &  -10.74$\pm$2.29  & -17.69$\pm$2.85  \\
    U-Net-50\%     & -7.05$\pm$0.97 & -19.40$\pm$2.91  &    -9.11$\pm$2.38  \\
    Step-learn-50\%@20 & -13.77$\pm$2.86      & -17.73$\pm$3.01  & -21.65$\pm$3.05 \\
    Diag-learn-50\%@20  & -13.85$\pm$2.71    & -17.50$\pm$2.97  & -21.35$\pm$3.05  \\
    \bottomrule
  \end{tabular}
\end{table}

\subsection{Image recovery with partial Fourier samples}

\begin{table} [!b]
  \caption{Averaged NMSE (dB) of methods trained with 50\%, tested on various partial Fourier cases.}
  \label{tbl:fourier}
  \centering \footnotesize
  \begin{tabular}{c | ccc}
    \toprule
    Method     & Test-30\%     & Test-50\% & Test-70\%  \\
    \midrule
    FISTA@100 & -20.46$\pm$3.49  &  -24.84$\pm$3.99  & -32.31$\pm$4.74 \\
    U-Net-50\%     & -18.22$\pm$3.43 & -20.17$\pm$4.08  &    -19.62$\pm$4.57   \\
    Step-learn-50\%@20 & -20.80$\pm$3.67      & -25.49$\pm$4.20  & -33.04$\pm$4.79  \\
    Diag-learn-50\%@20  & -20.80$\pm$3.65    & -25.52$\pm$4.17  & -33.06$\pm$4.78  \\
    \bottomrule
  \end{tabular}
\end{table}

\begin{figure} [!t]
	\centering
	\subfloat[\scriptsize Input images from 30, 50, and noisy 50\% partial Fourier samples] {
	\includegraphics[width=0.33\linewidth,trim={2cm 3cm 3cm 3cm},clip ]{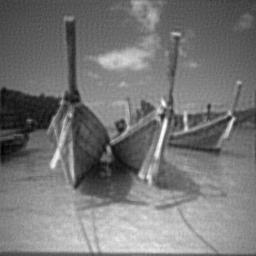}
	\includegraphics[width=0.33\linewidth,trim={2cm 3cm 3cm 3cm},clip ]{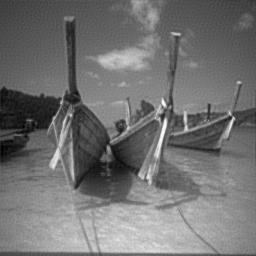}
	\includegraphics[width=0.33\linewidth,trim={2cm 3cm 3cm 3cm},clip ]{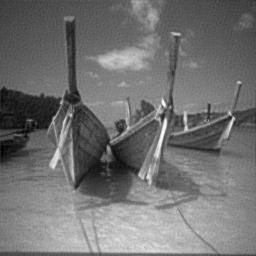}
	}
	\vspace{-0.5em} \vfil
	\subfloat[\scriptsize FISTA-b on 30, 50, noisy 50\% partial Fourier samples] { 
	\includegraphics[width=0.33\linewidth,trim={2cm 3cm 3cm 3cm},clip ]{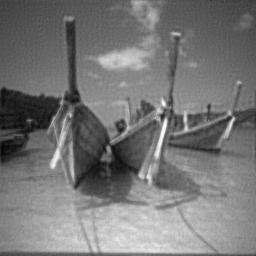}
	\includegraphics[width=0.33\linewidth,trim={2cm 3cm 3cm 3cm},clip ]{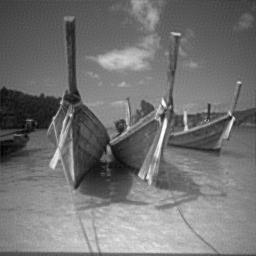}
	\includegraphics[width=0.33\linewidth,trim={2cm 3cm 3cm 3cm},clip ]{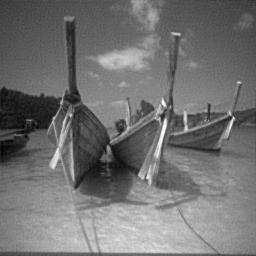}
	}
	\vspace{-0.5em} \vfil		
	\subfloat[\scriptsize U-Net trained with 50\%, tested on 30, 50, noisy 50\% partial Fourier samples] { 
	\includegraphics[width=0.33\linewidth,trim={2cm 3cm 3cm 3cm},clip ]{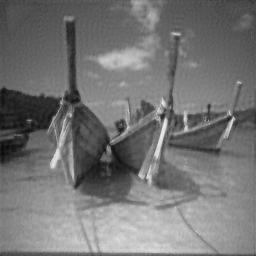}
	\includegraphics[width=0.33\linewidth,trim={2cm 3cm 3cm 3cm},clip ]{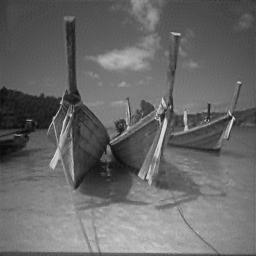}
	\includegraphics[width=0.33\linewidth,trim={2cm 3cm 3cm 3cm},clip ]{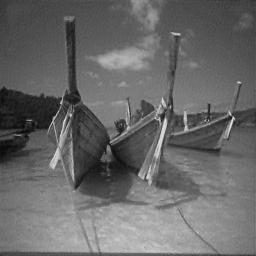}
	}
	\vspace{-0.5em} \vfil		
	\subfloat[\scriptsize Step-learned SGP trained with 50\%, tested on 30, 50, noisy 50\%] { 
	\includegraphics[width=0.33\linewidth,trim={2cm 3cm 3cm 3cm},clip ]{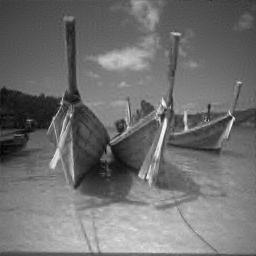}
	\includegraphics[width=0.33\linewidth,trim={2cm 3cm 3cm 3cm},clip ]{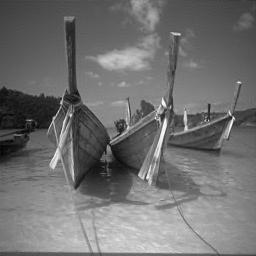}
	\includegraphics[width=0.33\linewidth,trim={2cm 3cm 3cm 3cm},clip ]{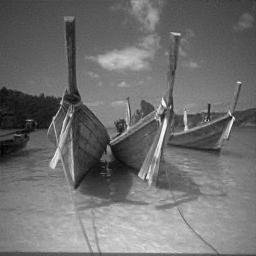}
	}
	\vspace{-0.5em} \vfil		
	\subfloat[\scriptsize Diag-learned SGP trained with 50\%, tested on 30, 50, noisy 50\%] { 
	\includegraphics[width=0.33\linewidth,trim={2cm 3cm 3cm 3cm},clip ]{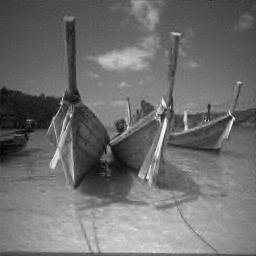}
	\includegraphics[width=0.33\linewidth,trim={2cm 3cm 3cm 3cm},clip ]{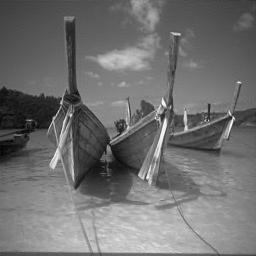}
	\includegraphics[width=0.33\linewidth,trim={2cm 3cm 3cm 3cm},clip ]{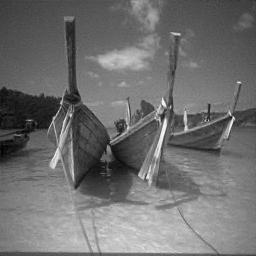}
	}
	\vspace{0.5em} \vfil		
	\caption{Recovered images from partial Fourier sampling. DNNs were trained with 50\% and tested on 30, 50, noisy 50\% 
	samples. 40 iterations were run for FISTA-b and our proposed methods.}
	\label{fig:fourier}
		\vspace{-1em}
\end{figure}

\begin{figure} [!t]
	\centering
	\subfloat[\scriptsize Train-50\%,Test-30\%] {\includegraphics[width=0.49\columnwidth]{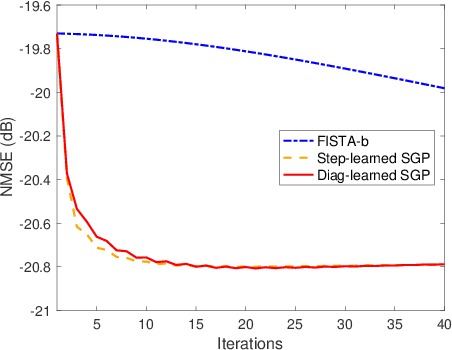}}
	\subfloat[\scriptsize Train-50\%,Test-50\%] {\includegraphics[width=0.49\columnwidth]{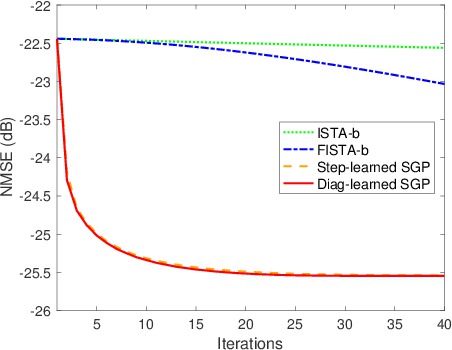}}
		\vspace{-0.5em}	\vfil
	\subfloat[\scriptsize Train-50\%,Test-50\% noise] {\includegraphics[width=0.49\columnwidth]{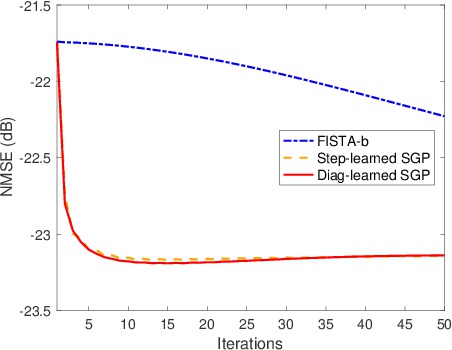}}
	\subfloat[\scriptsize Train-50\%,Test-70\%] {\includegraphics[width=0.49\columnwidth]{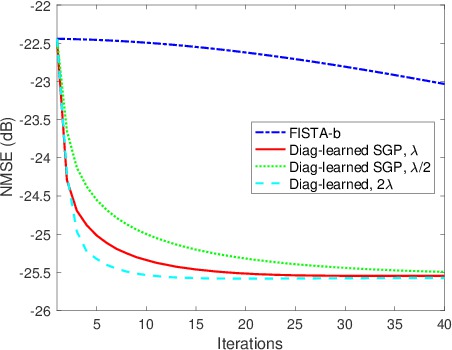}}
\vspace{0.5em}	
	\caption{NMSE over iterations for all methods in partial Fourier recovery trained on 50\% sampling, tested on 30 (a), 50 (b), noisy 50\% (c) samplings or with $\lambda/2$, $2\lambda$ (d).}
	\label{fig:fourier2}
			\vspace{-1em}
\end{figure}

Similar simulations were performed for image recovery with partial Fourier samples.
Note that the input image of the DNN has four channels such that
the first two channels are the real and imaginary of the estimated image.
Initial images in Figure~\ref{fig:fourier} were obtained using inverse Fourier transform with zero padding.
All results are reported in Figures~\ref{fig:fourier},~\ref{fig:fourier2} and Tables~\ref{tbl:fourier},~\ref{tbl:fourier2}.

For all test cases, non-iterative U-Net yielded the worse results among all methods including FISTA at 100 iteration. 
Forward models for partial Fourier sampling is much more complicated than forward models for inpainting, and thus
more complicated DNN with much more dataset seems desirable for better performance.
Our proposed methods without re-training yielded robust and excellent performance at early iteration for all test cases including
the same sampling rate (50\%), different sampling rates (30, 70\%), and additional measurement noise case (noisy 50\%).
Thus, our proposed DNNs does seem robust to different models and noise in compressive sensing recovery with partial Fourier samples
that was inherited from conventional optimization based algorithms.

\begin{table} [!b]
  \caption{Averaged NMSE (dB) of methods trained with 50\%, tested on various partial Fourier cases.}
  \label{tbl:fourier2}
  \centering \footnotesize
  \begin{tabular}{c | cc}
    \toprule
    Method     &  Test-50\% &  Test-50\% noisy \\
    \midrule
    FISTA@100  &  -24.84$\pm$3.99  &  -22.79$\pm$2.85 \\
    U-Net-50\%     &  -20.17$\pm$4.08  &       -19.85$\pm$3.79  \\
    Step-learn-50\%@20     & -25.49$\pm$4.20  & -23.16$\pm$2.78 \\
    Diag-learn-50\%@20   & -25.52$\pm$4.17  & -23.18$\pm$2.76 \\
    \bottomrule
  \end{tabular}
\end{table}

\subsection{Robustness to regularization parameters}

We investigate the robustness of our proposed methods by running them that were trained with the original regularization parameter on the test set with
different regularization parameters that are half of the original value and twice (2x) of the original value. 

Figures~\ref{fig:inpaint_robust} (d),~\ref{fig:fourier2} (d) illustrate that our proposed methods were robust to small changes in regularization parameters such as half or twice.
However, large changes such as 10 times smaller or larger than the original parameter seem to break fast empirical convergence properties of our proposed methods.
These phenomena were expected since changing regularization parameters leads to changing ground truth images, thus our DNNs whose inputs are dependent on
current estimate and its corresponding gradient should behave in a different way ($e.g.$, if the current estimate is the same as the converged solution with original regularization parameter, then, zero stepsize should be obtained for the problem with the same regularization parameter, but non-zero stepsize should be obtained for different problem with different regularization parameter). 
Thus, large changes in regularization parameter may require re-training the DNN.

\subsection{Image deblurring}

\begin{figure} [!b]
	\centering
			\vspace{-1em}
	\subfloat[\scriptsize Blurred image] {\includegraphics[width=0.32\columnwidth,trim={0 0 0 1cm},clip ]{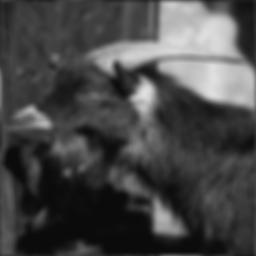}}
	\subfloat[\scriptsize FISTA-b] { \includegraphics[width=0.32\columnwidth,trim={0 0 0 1cm},clip ]{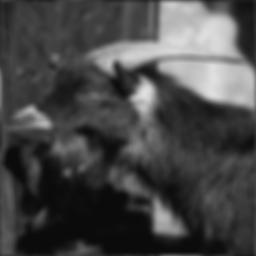}}
	\subfloat[\scriptsize Step-learned SGP] { \includegraphics[width=0.32\columnwidth,trim={0 0 0 1cm},clip ]{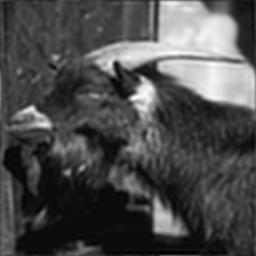}}
	\vfil			\vspace{-0.5em}  
	\subfloat[\scriptsize Diag-learned SGP] { \includegraphics[width=0.32\columnwidth,trim={0 0 0 1cm},clip ]{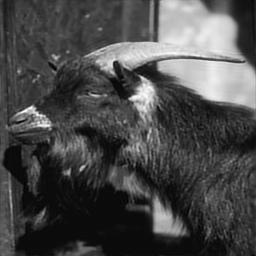}}
	\subfloat[\scriptsize Iteration vs. NMSE (dB)] {\includegraphics[width=0.48\columnwidth]{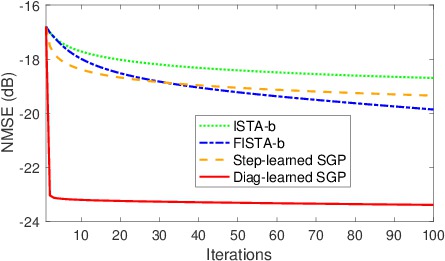}}
			\vspace{0.5em}  
	\caption{Reconstructed images for deblurring from (a) input blurred image using Gaussian kernel with $\sigma=2$: (b) FISTA with backtracking, (c)  step-learned SGP, (d)  diag-learned SGP, all at the 10th iteration. Proposed methods yielded faster initial convergence rates than FISTA-b (e) average convergence for 50 images.}
	\label{fig:deblur}
\end{figure}

Proposed methods were applied to image deblurring problems. Images were blurred using Gaussian kernel with $\sigma=2$. Then, image deblurring was performed with the regularization parameter 0.00001. Note that the initial data fidelity term for deblurring problem is usually much larger than other inverse problems such as inpainting problems.
Unlike other inverse problems in image processing, learned diagonal matrix based relaxed SGP yielded the best image quality among all compared methods as shown in Figure~\ref{fig:deblur} qualitatively and quantitatively. It seems that large discrepancy in the data fidelity term was quickly compensated when using the learned diagonal matrix in SGP.

\subsection{Sparse-view medical image reconstruction}

\begin{figure} [!t]
	\centering
	\subfloat[\scriptsize Input image from 144 views] {\includegraphics[width=0.48\columnwidth,trim={0 1cm 0 5cm},clip ]{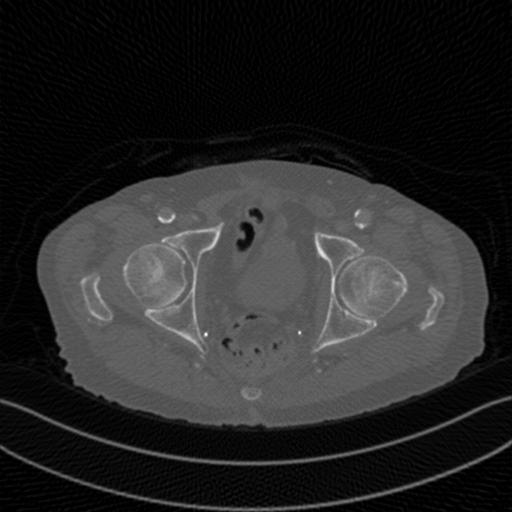}}
	\subfloat[\scriptsize FISTA-b] {\includegraphics[width=0.48\columnwidth,trim={0 1cm 0  5cm},clip ]{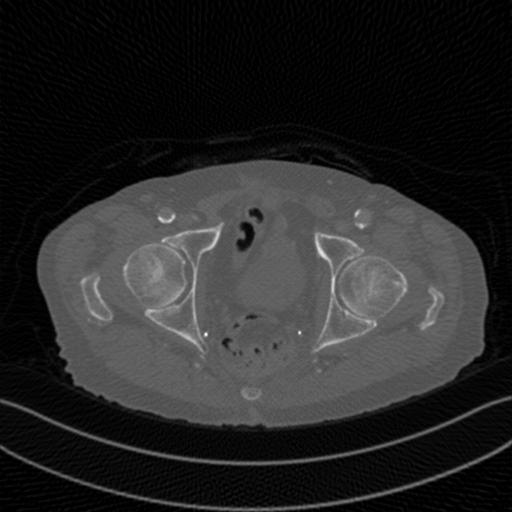}}
	\vfil 	\vspace{-.5em}
	\subfloat[\scriptsize Diag-learned SGP] {\includegraphics[width=0.48\columnwidth,trim={0 1cm 0 5cm},clip ]{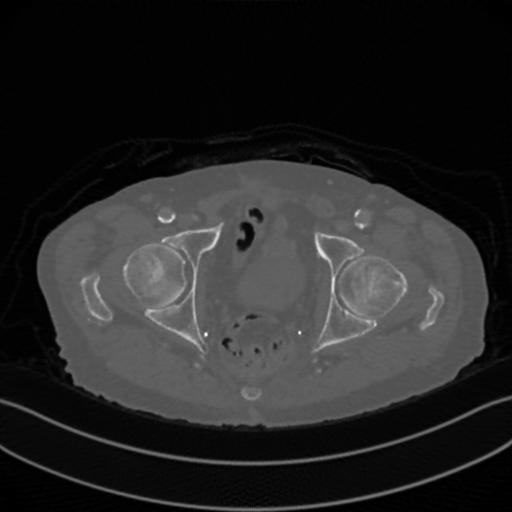}}
	\subfloat[\scriptsize Iteration vs. NMSE (dB)] {\includegraphics[width=0.48\columnwidth,trim={0 0 0  1cm},clip]{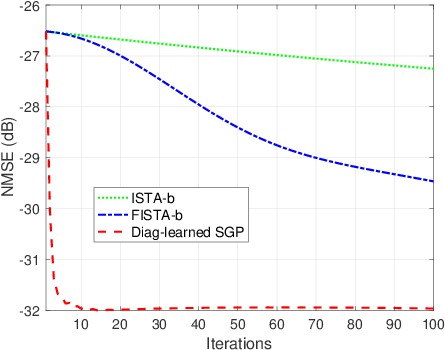}}
	\vspace{.5em}
	\caption{Reconstructed images for sparse-view CT from (a) Input image (144 views) : (b) FISTA with backtracking, 
	(c) Proposed diag-learned SGP, all at the 10th iteration. 
	(d) Proposed method yielded faster convergence rate than FISTA-b.}
	\label{fig:med}
				\vspace{-1em}
\end{figure}

Lastly, our proposed method was investigated for sparse-view CT image reconstruction.  Initial image in Figure~\ref{fig:med} (a) was obtained by filtered back-projection from 144 views of projections and had streaking artifacts. With the regularization parameter 0.0005, we ran FISTA-b and proposed diag-learned SGP. At the 10th iteration, our proposed method
yielded visually better image than FISTA as illustrated in Figure~\ref{fig:med} (b) and (c). Figure~\ref{fig:med} (d) shows that our proposed method achieved faster convergence rate than FISTA-b.

\subsection{Limited robustness to other measurements}

Robustness of our trained DNN for determining stepsize or diagonal matrix when different forward models are used.
Our proposed methods have shown robustness for the problems such as image inpainting.
It shows that the trained DNN still achieved much faster convergence than FISTA. 
Similar tendency was observed for image inpainting with 70\% sampling.
The trained DNN also yielded robust performance for lower or higher (70\%) sampling in partial Fourier image recovery.
For image deblurring problems with different blur levels, the trained DNN yielded sub-optimal performance compared to FISTA. However, note that step-learned SGP yielded relatively robust performance to diag-learned SGP and it yielded better performance than FISTA for early iterations. For sparse-view image reconstruction, the trained DNN
with 144 views did not yield good performance for the test with 45 views. Thus, the robustness of the trained DNN to other forward models seems application-dependent. However, many DNN based algorithms are not robust to other measurement models~\cite{Jin:2017iz}.

\section{Conclusion}

We proposed a new way of using CNNs for empirically accelerating convergence 
for inverse problems in image processing with dynamic parameter selections over iterations with different forward models and 
without breaking theoretical properties such as convergence and robustness.
Our trained DNN enabled SGP to empirically outperform FISTA
that is theoretically faster than SGP and 
yield robust performance compared to direct mapping DNNs. % without re-training.

	\section*{Acknowledgments}
	
	This work was supported partly by 
	Basic Science Research Program through the National Research Foundation of Korea(NRF) 
	funded by the Ministry of Education(NRF-2017R1D1A1B05035810),
	the Technology Innovation Program or Industrial Strategic Technology Development Program 
	(10077533, Development of robotic manipulation algorithm for grasping/assembling 
	with the machine learning using visual and tactile sensing information) 
	funded by the Ministry of Trade, Industry \& Energy (MOTIE, Korea), and a grant of the Korea Health Technology R\&D Project 
	through the Korea Health Industry Development Institute (KHIDI), 
	funded by the Ministry of Health \& Welfare, Republic of Korea (grant number: HI18C0316).

{\small
\bibliographystyle{ieee_fullname}
\bibliography{arxiv_optstep_v1}

\begin{thebibliography}{10}\itemsep=-1pt

\bibitem{andrychowicz2016learning}
Marcin Andrychowicz, Misha Denil, Sergio Gomez, Matthew~W Hoffman, David Pfau,
  Tom Schaul, Brendan Shillingford, and Nando De~Freitas.
\newblock Learning to learn by gradient descent by gradient descent.
\newblock In {\em Advances in neural information processing systems}, pages
  3981--3989, 2016.

\bibitem{amfm_pami2011}
Pablo Arbelaez, Michael Maire, Charless Fowlkes, and Jitendra Malik.
\newblock Contour detection and hierarchical image segmentation.
\newblock {\em IEEE Transactions on Pattern Analysis and Machine Intelligence},
  33(5):898--916, May 2011.

\bibitem{Armijo:1966hc}
Larry Armijo.
\newblock {Minimization of functions having lipschitz continuous first partial
  derivatives}.
\newblock {\em Pacific Journal of Mathematics}, 16(1):1--3, Jan. 1966.

\bibitem{Beck:2009tk}
Amir Beck and Marc Teboulle.
\newblock {A fast iterative shrinkage-thresholding algorithm for linear inverse
  problems}.
\newblock {\em SIAM Journal on Imaging Sciences}, 2(1):183--202, 2009.

\bibitem{Bonettini:2015hu}
S Bonettini and M Prato.
\newblock {New convergence results for the scaled gradient projection method}.
\newblock {\em Inverse Problems}, 31(9):095008, Sept. 2015.

\bibitem{Bonettini:2008ud}
Silvia Bonettini, Riccardo Zanella, and Luca Zanni.
\newblock {A scaled gradient projection method for constrained image
  deblurring}.
\newblock {\em Inverse Problems}, 25(1):015002, 2008.

\bibitem{boyd2011distributed}
Stephen Boyd, Neal Parikh, Eric Chu, Borja Peleato, Jonathan Eckstein, et~al.
\newblock Distributed optimization and statistical learning via the alternating
  direction method of multipliers.
\newblock {\em Foundations and Trends in Machine learning}, 3(1):1--122, 2011.

\bibitem{Candes:2006eq}
E~J Candes, J Romberg, and T Tao.
\newblock {Robust uncertainty principles: exact signal reconstruction from
  highly incomplete frequency information}.
\newblock {\em IEEE Transactions on Information Theory}, 52(2):489--509, Jan.
  2006.

\bibitem{Chang:2017fz}
J~H~Rick Chang, Chun-Liang Li, Barnabas Poczos, and B~V K~Vijaya Kumar.
\newblock One network to solve them all - solving linear inverse problems using
  deep projection models.
\newblock In {\em IEEE International Conference on Computer Vision (ICCV)},
  pages 5889--5898, 2017.

\bibitem{Chen:2018vc}
Xiaohan Chen, Jialin Liu, Zhangyang Wang, and Wotao Yin.
\newblock {Theoretical Linear Convergence of Unfolded ISTA and Its Practical
  Weights and Thresholds}.
\newblock In {\em Advances in Neural Information Processing Systems (NeurIPS)},
  pages 9061--9071, 2018.

\bibitem{donoho2009message}
David~L Donoho, Arian Maleki, and Andrea Montanari.
\newblock Message-passing algorithms for compressed sensing.
\newblock {\em Proceedings of the National Academy of Sciences},
  106(45):18914--18919, 2009.

\bibitem{figueiredo2003algorithm}
M{\'a}rio~AT Figueiredo and Robert~D Nowak.
\newblock {An EM algorithm for wavelet-based image restoration}.
\newblock {\em IEEE Transactions on Image Processing}, 12(8):906--916, 2003.

\bibitem{Giryes:2018by}
Raja Giryes, Yonina~C Eldar, Alex~M Bronstein, and Guillermo Sapiro.
\newblock {Tradeoffs Between Convergence Speed and Reconstruction Accuracy in
  Inverse Problems}.
\newblock {\em IEEE Transactions on Signal Processing}, 66(7):1676--1690, Feb.
  2018.

\bibitem{gregor2010learning}
Karol Gregor and Yann LeCun.
\newblock Learning fast approximations of sparse coding.
\newblock In {\em International Conference on Machine Learning (ICML)}, pages
  399--406, 2010.

\bibitem{gupta2018tmi}
Harshit Gupta, Kyong~Hwan Jin, Ha~Q Nguyen, Michael~T McCann, and Michael
  Unser.
\newblock {CNN-based projected gradient descent for consistent CT image
  reconstruction}.
\newblock {\em IEEE Transactions on Medical Imaging}, 37(6):1440--1453, 2018.

\bibitem{he19tmi}
J. {He}, Y. {Yang}, Y. {Wang}, D. {Zeng}, Z. {Bian}, H. {Zhang}, J. {Sun}, Z.
  {Xu}, and J. {Ma}.
\newblock Optimizing a parameterized plug-and-play admm for iterative low-dose
  ct reconstruction.
\newblock {\em IEEE Transactions on Medical Imaging}, 38(2):371--382, Feb 2019.

\bibitem{Jin:2017iz}
Kyong~Hwan Jin, Michael~T McCann, Emmanuel Froustey, and Michael Unser.
\newblock {Deep Convolutional Neural Network for Inverse Problems in Imaging}.
\newblock {\em IEEE Transactions on Image Processing}, 26(9):4509--4522, Sept.
  2017.

\bibitem{Kulkarni:2016jea}
Kuldeep Kulkarni, Suhas Lohit, Pavan Turaga, Ronan Kerviche, and Amit Ashok.
\newblock {ReconNet: Non-iterative reconstruction of images from compressively
  sensed measurements}.
\newblock In {\em IEEE Conference on Computer Vision and Pattern Recognition
  (CVPR)}, pages 449--458, 2016.

\bibitem{noise2noise}
Jaakko Lehtinen, Jacob Munkberg, Jon Hasselgren, Samuli Laine, Tero Karras,
  Miika Aittala, and Timo Aila.
\newblock {N}oise2{N}oise: Learning image restoration without clean data.
\newblock In {\em International Conference on Machine Learning (ICML)}, pages
  2965--2974, 2018.

\bibitem{Li:2017lto}
Ke Li and Jitendra Malik.
\newblock Learning to optimize.
\newblock In {\em International Conference on Learning Representations (ICLR)},
  2017.

\bibitem{Lim:2017it}
Bee Lim, Sanghyun Son, Heewon Kim, Seungjun Nah, and Kyoung~Mu Lee.
\newblock {Enhanced Deep Residual Networks for Single Image Super-Resolution}.
\newblock In {\em IEEE Conference on Computer Vision and Pattern Recognition
  Workshops (CVPRW)}, pages 1132--1140, 2017.

\bibitem{Liu:2019ta}
Jialin Liu, Xiaohan Chen, Zhangyang Wang, and Wotao Yin.
\newblock {ALISTA: Analytic Weights Are As Good As Learned Weights in LISTA}.
\newblock In {\em International Conference on Learning Representations (ICLR)},
  2019.

\bibitem{Mairal:2009gl}
Julien Mairal, Francis~R Bach, Jean Ponce, Guillermo Sapiro, and Andrew
  Zisserman.
\newblock {Non-local sparse models for image restoration}.
\newblock In {\em IEEE International Conference on Computer Vision (ICCV)},
  pages 2272--2279, 2009.

\bibitem{metzler2017learned}
Chris Metzler, Ali Mousavi, and Richard Baraniuk.
\newblock {Learned D-AMP: Principled neural network based compressive image
  recovery}.
\newblock In {\em Advances in Neural Information Processing Systems (NIPS)},
  pages 1772--1783, 2017.

\bibitem{Moreau:2017wj}
Thomas Moreau and Joan Bruna.
\newblock {Understanding Trainable Sparse Coding via Matrix Factorization}.
\newblock In {\em International Conference on Learning Representations (ICLR)},
  2017.

\bibitem{Patel:2012gz}
V~M Patel, R Maleh, A~C Gilbert, and R Chellappa.
\newblock Gradient-based image recovery methods from incomplete {F}ourier
  measurements.
\newblock {\em IEEE Transactions on Image Processing}, 21(1):94--105, Jan.
  2012.

\bibitem{Ronneberger:2015vw}
O Ronneberger, P Fischer, and T Brox.
\newblock {U-Net: Convolutional Networks for Biomedical Image Segmentation}.
\newblock In {\em Medical Image Computing and Computer-Assisted Intervention
  (MICCAI)}, pages 234--241, 2015.

\bibitem{Roth:2005hu}
Stefan Roth and Michael~J Black.
\newblock Fields of experts: A framework for learning image priors.
\newblock In {\em IEEE Conference on Computer Vision and Pattern Recognition
  (CVPR)}, pages 860--867, 2005.

\bibitem{ryu19a}
Ernest Ryu, Jialin Liu, Sicheng Wang, Xiaohan Chen, Zhangyang Wang, and Wotao
  Yin.
\newblock Plug-and-play methods provably converge with properly trained
  denoisers.
\newblock In {\em International Conference on Machine Learning}, pages
  5546--57, 2019.

\bibitem{sun2016deep}
Jian Sun, Huibin Li, Zongben Xu, et~al.
\newblock {Deep ADMM-Net for compressive sensing MRI}.
\newblock In {\em Advances in Neural Information Processing Systems (NIPS)},
  pages 10--18, 2016.

\bibitem{tirer2018image}
Tom Tirer and Raja Giryes.
\newblock Image restoration by iterative denoising and backward projections.
\newblock {\em IEEE Transactions on Image Processing}, 28(3):1220--1234, 2018.

\bibitem{xie2012image}
Junyuan Xie, Linli Xu, and Enhong Chen.
\newblock Image denoising and inpainting with deep neural networks.
\newblock In {\em Advances in Neural Information Processing Systems (NIPS)},
  pages 341--349, 2012.

\bibitem{Zhang:2018wz}
Jian Zhang and Bernard Ghanem.
\newblock {ISTA-Net: Interpretable Optimization-Inspired Deep Network for Image
  Compressive Sensing}.
\newblock In {\em IEEE Conference on Computer Vision and Pattern Recognition
  (CVPR)}, pages 1828--1837, 2018.

\bibitem{zhang2017beyond}
Kai Zhang, Wangmeng Zuo, Yunjin Chen, Deyu Meng, and Lei Zhang.
\newblock {Beyond a Gaussian denoiser: Residual learning of deep CNN for image
  denoising}.
\newblock {\em IEEE Transactions on Image Processing}, 26(7):3142--3155, 2017.

\bibitem{Zhong:2013in}
Lin Zhong, Sunghyun Cho, Dimitris Metaxas, Sylvain Paris, and Jue Wang.
\newblock Handling noise in single image deblurring using directional filters.
\newblock In {\em IEEE Conference on Computer Vision and Pattern Recognition
  (CVPR)}, pages 612--619, 2013.

\bibitem{Zoran:2011jn}
Daniel Zoran and Yair Weiss.
\newblock {From learning models of natural image patches to whole image
  restoration}.
\newblock In {\em IEEE International Conference on Computer Vision (ICCV)},
  pages 479--486, 2011.

\end{thebibliography}
}

\end{document}